\documentclass[conference]{IEEEtran}
\IEEEoverridecommandlockouts

\usepackage[round,authoryear]{natbib}
\AtBeginDocument{%
  \bibpunct{(}{)}{;}{a}{ }{,}
  \setcitestyle{authoryear,round,aysep={ },yysep={;}}%
  \renewcommand{\cite}{\citep}%
  \renewcommand{\citet}{\citep}
}
\usepackage{amsmath,amssymb,amsfonts}
\usepackage{algorithmic}
\usepackage{graphicx}
\usepackage[export]{adjustbox}
\usepackage{textcomp}
\usepackage{xcolor}
\usepackage{booktabs}
\usepackage{array}
\usepackage{multirow}
\usepackage{url}
\usepackage{balance}
\usepackage{dblfloatfix} 
\usepackage{cuted}      
\usepackage{caption}
\usepackage{placeins}   

\makeatletter
\def\fnum@table{Table~\Roman{table}}
\makeatother
\AtBeginDocument{%
  \setlength{\parskip}{0.08\baselineskip}%
  \raggedbottom
}

\makeatletter
\g@addto@macro\normalsize{%
  \abovedisplayskip 6pt\relax
  \belowdisplayskip 6pt\relax
  \abovedisplayshortskip 6pt\relax
  \belowdisplayshortskip 6pt\relax
}
\makeatother

\makeatletter
\renewcommand{\section}{\@startsection{section}{1}{\z@}%
  {1.5ex plus 1ex minus .2ex}%
  {0.25ex plus .1ex}%
  {\normalfont\normalsize\centering\scshape}}
\renewcommand{\subsection}{\@startsection{subsection}{2}{\z@}%
  {1.4ex plus 0.5ex minus .2ex}%
  {0.25ex plus .1ex}%
  {\normalfont\normalsize\itshape}}
\renewcommand{\paragraph}{\@startsection{paragraph}{4}{\z@}%
  {0.6ex plus 0.2ex}%
  {-0.15em}
  {\normalfont\normalsize\itshape}}
\makeatother

\def\BibTeX{{\rm B\kern-.05em{\sc i\kern-.025em b}\kern-.08em
    T\kern-.1667em\lower.7ex\hbox{E}\kern-.125emX}}

\makeatletter
\let\reva@origmaketitle\maketitle
\renewcommand{\maketitle}{%
  \begingroup
  \let\LARGE\large
  \reva@origmaketitle
  \endgroup
}
\makeatother

\begin{document}

\title{ReVA: A Region-Aware Visual Assistant for Visually Grounded Question Answering}

\author{
\IEEEauthorblockN{Anoop Senthil\IEEEauthorrefmark{1}}
\IEEEauthorblockA{\textit{School of Electronic Engineering and Computer Science}\\
\textit{Queen Mary University of London}\\
London, United Kingdom\\
a.senthil@se25.qmul.ac.uk}
\thanks{\IEEEauthorrefmark{1}The source code and pre-trained weights for ReVA are publicly available at {\color{blue}\texttt{https://github.com/anoop675/reva}}.}
}

\maketitle

\begin{abstract}
Multimodal Large Language Models (MLLMs) have achieved remarkable progress in Visual Question Answering (VQA), yet they continue to struggle with questions requiring precise spatial reasoning and fine-grained visual understanding.
These limitations often manifest as object, attribute, and spatial hallucinations, where models generate confident but visually unsupported responses due to insufficient region-level and fine-grained visual grounding.
To address this challenge, we propose ReVA, a region-aware VQA model that employs a frozen CLIP ViT-L/14 Vision Transformer (ViT) and a Qwen2.5-7B-Instruct large language model (LLM) connected through a dual bridge that aligns both whole-image and region-level representations with the LLM's embedding space.
The image bridge maps final transformer block features into image tokens.
The region bridge maps cropped features from enriched intermediate features across ViT blocks so early texture and later object cues are more evident, into $K$ region tokens for every bounding box.
ReVA uses a detector stack that supplies automatic zero-shot bounding boxes that are both question-agnostic and question-dependent, using RAM++ (Recognize Anything Model), spaCy, and Grounding~DINO.
The image tokens and region tokens are concatenated as an LLM prompt prefix to jointly encode scene-level context and fine-grained regional evidence when answering questions.
Evaluated on VQAv2, MMBench, POPE, and SEED-Bench, ReVA achieves 82.85\% mean F1 on POPE, compared with 81.14\% for an image-token baseline without region tokens.
These results demonstrate that explicit region-aware visual representations reduce object hallucination and improve the factual grounding of MLLMs.
\end{abstract}


\section{Introduction}
\label{sec:introduction}

This work presents \textbf{ReVA} (Region-Aware Visual Assistant), a region-aware VQA model built to ground language in both the broader scene and the fine-grained visual evidence within it (Fig.~\ref{fig:architecture-overview}).
At its core, ReVA pairs a frozen CLIP ViT-L/14 Vision Transformer (ViT) \citep{radford2021learning,dosovitskiy2020vit}, operating at 336\,px, with a Qwen2.5-7B-Instruct backbone \citep{liu2024qwen25} adapted via LoRA \citep{hu2022lora}, linked by an image bridge and a region bridge.
The image bridge maps final transformer block features into 576 image tokens.
For local evidence, RAM++ (Recognize Anything Model)~\citep{huang2023rampp} tags objects in the image, spaCy \citep{honnibal2020spacy} adds nouns from the question so named objects are not missed, and Grounding~DINO \citep{ren2024grounding} localises each tag to a box.
This makes the region detector's proposals both question-agnostic and question-dependent.
The region bridge then crops intermediate ViT-block features with those boxes and maps each crop to region tokens.
The LLM therefore sees what lies inside each box, and the boxes themselves never appear in the prompt as text.
Training has three stages so each bridge aligns before the LLM combines them: Stage~1 trains only the image bridge on whole-image captions (CLIP and Qwen frozen); Stage~2 trains the region bridge on box-linked descriptions with frozen backbones; Stage~3 adapts Qwen with LoRA to answer from the concatenated image and region tokens.
In later comparisons, the \emph{image-token} baseline uses the Stage~1 image bridge plus a separately trained Stage~3 LoRA with no region tokens; \emph{ReVA} is the full model after all three stages, using region tokens from proposed boxes at inference.

Recent MLLMs---LLaVA \citep{liu2023llava}, InstructBLIP \citep{dai2023instructblip}, Qwen-VL \citep{bai2023qwenvl}---encode the image, project patches into the LLM, and generate an answer.
That whole-image path is strong for scene-level understanding, yet even a dense 576-token grid at 336\,px can leave spatial relations and attribute--instance binding ambiguous without explicit local evidence.
GPT4RoI~\citep{zhang2023gpt4roi} supplies learned RoI tokens; Shikra~\citep{chen2023shikra} supplies box text.
ReVA follows the RoI-token approach, expands each box into $K$ region tokens rather than a single pooled vector, and keeps the user question as ordinary natural language.
Local evidence enters only as region tokens, as a prefix of the LLM prompt.

\section{Related Work}
\label{sec:related}

\paragraph{Connecting vision encoders to language models}
The dominant recipe for modern MLLMs starts from a pretrained vision encoder and a pretrained LLM, learns a bridge between them, and often adapts the LLM while keeping the vision encoder frozen.
BLIP-2~\citep{li2022blip2} introduced a lightweight Querying Transformer (Q-Former) that extracts a small set of query embeddings from a frozen image encoder for the LLM to read, and InstructBLIP~\citep{dai2023instructblip} made this bridge instruction-aware by also feeding the task instruction into the Q-Former, so the visual features it extracts depend on the question being asked.
LLaVA~\citep{liu2024visual} showed that an even simpler bridge---a linear projection, upgraded to a two-layer MLP in LLaVA-1.5~\citep{liu2023llava}---trained on visual-instruction data is remarkably effective, while Qwen-VL~\citep{bai2023qwenvl} compressed patch features into a fixed-length sequence through a position-aware cross-attention adapter.
ReVA's image bridge sits in this lineage: a frozen CLIP encoder~\citep{radford2021learning} feeds a two-layer MLP (the Image Feature Projector) into Qwen2.5-7B-Instruct~\citep{liu2024qwen25}.
On its own, that image bridge presents the image as one holistic token set---strong for scene-level understanding, but with no explicit handle on individual objects.

\paragraph{Making MLLMs region-aware}
Two broad strategies have emerged to restore that missing locality.
The first encodes location in the text stream: Shikra~\citep{chen2023shikra} writes bounding boxes as plain numbers in natural language with no extra vocabulary or detector, Kosmos-2~\citep{peng2023kosmos2} attaches discrete location tokens to phrases in a Markdown-style hyperlink format (trained on its large-scale GRIT corpus), and Qwen-VL~\citep{bai2023qwenvl} wraps normalised box strings in special tokens.
These schemes are elegant, but they ask the LLM to handle spatial position as text or location tokens rather than as cropped visual evidence.
The second strategy instead feeds the LLM visual features taken from the region itself.
GPT4RoI~\citep{zhang2023gpt4roi} builds a multi-level CLIP feature pyramid, incorporates feature coordinates~\citep{liu2018coordconv} for absolute position, and uses RoI Align~\citep{he2017maskrcnn} so that each box replaces a \texttt{<region>} placeholder with region features interleaved in the instruction.
Ferret~\citep{you2024ferret,zhang2024ferretv2} takes a hybrid route: discrete coordinates plus continuous features from a spatial-aware visual sampler that covers points, boxes, and free-form shapes.
GLaMM~\citep{rasheed2024glamm} pushes grounding down to pixel-level segmentation masks, and Groma~\citep{ma2024groma} treats regions as first-class visual tokens---encoding proposed regions into region tokens alongside global tokens so the LLM can ground by referring to those tokens rather than regressing coordinates.
ReVA adopts the region-feature-as-token philosophy for VQA.
Where it differs from GPT4RoI is how a box becomes tokens: rather than fusing the pyramid levels into a single RoI embedding, ReVA mixes them with multi-head self-attention and expands each box into $K$ tokens through multi-token attention pooling, giving Qwen a richer local descriptor beside the 576 image tokens.
It also keeps the user question as plain natural language without interleaving boxes, location tokens, or \texttt{<region>} placeholders into the query text.

\paragraph{Where the regions come from}
At inference, GPT4RoI accepts boxes from the user or from an off-the-shelf detector; Ferret's referring path takes user-specified regions (points, boxes, or free-form shapes) while its grounding path emits boxes in the response; and Groma trains its own region proposer.
ReVA instead builds the region set automatically from off-the-shelf open-vocabulary detectors: RAM++ (Recognize Anything Model)~\citep{huang2023rampp} proposes open-set tags for what is present, and Grounding~DINO~\citep{ren2024grounding} localises those tags into boxes from a text prompt.
No detector is retrained, and the boxes only decide where to crop features rather than entering the prompt as text.
This design tests whether region-aware gains are available under standard VQA inputs that use ordinary natural language questions with no additional spatial context interleaved in the question itself.

\section{System Architecture}
\label{sec:architecture}

\begin{figure*}[!t]
\centering
\includegraphics[width=0.80\textwidth,keepaspectratio]{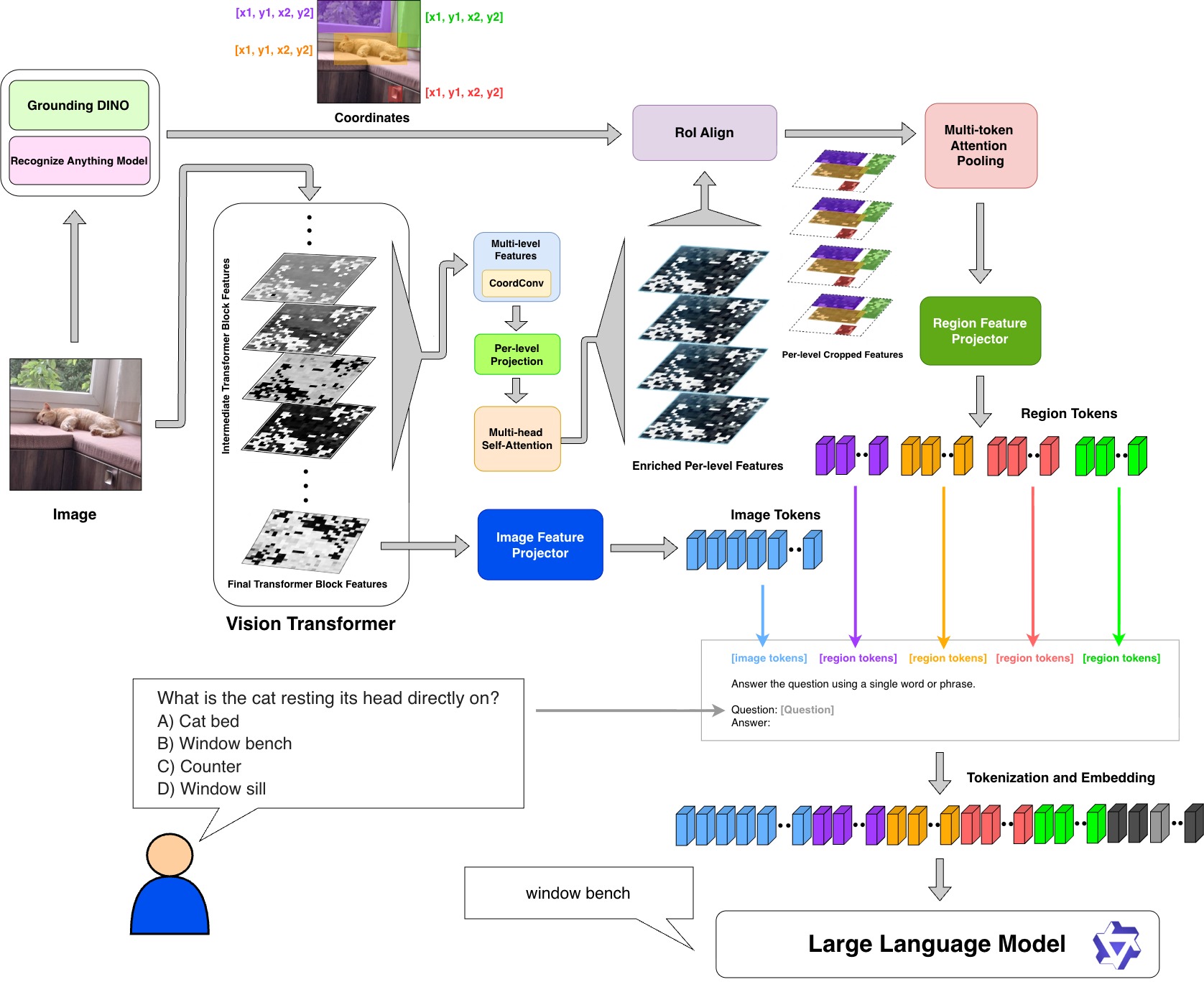}
\caption{Overview of ReVA. The Vision Transformer is CLIP ViT-L/14. Image bridge: final transformer block features pass through the Image Feature Projector to image tokens. Region bridge: intermediate transformer block features are enriched (CoordConv, per-level projection, multi-head self-attention), cropped with RoI Align using RAM++ (Recognize Anything Model) tags localised by Grounding~DINO, then pooled and mapped by the Region Feature Projector to region tokens.}
\label{fig:architecture-overview}
\end{figure*}

Fig.~\ref{fig:architecture-overview} shows ReVA's dual-bridge pipeline that supplies both image tokens and region tokens into Qwen.
The image is encoded once by CLIP ViT-L/14, the Vision Transformer in the figure.
The image bridge passes final transformer block features through the Image Feature Projector as 576 image tokens.
The region bridge enriches intermediate transformer block features with CoordConv, per-level projection, and multi-head self-attention, then crops them with proposed box coordinates; they become $K$ region tokens per box.
Those visual tokens are concatenated with a short text footer (answer format, question, ``Answer:''); Qwen then generates the reply.
Stages~1--2 train the image bridge and the region bridge with frozen CLIP and Qwen; Stage~3 adds LoRA \citep{hu2022lora} in Qwen.

\subsection{Image Alignment (image bridge)}

The image bridge is a LLaVA-style two-layer MLP---the Image Feature Projector---that maps each CLIP patch into Qwen's embedding space.
Images are resized to $336{\times}336$ and encoded by CLIP ViT-L/14 \citep{radford2021learning} into a $24{\times}24$ grid of 1024-d patch tokens (CLS discarded).
Let $\mathbf{f}^{\mathrm{patch}}\in\mathbb{R}^{1024}$ denote one such CLIP patch token.
The Image Feature Projector maps each patch into Qwen's 3584-d space, yielding an image token $\mathbf{z}^{\mathrm{image}}\in\mathbb{R}^{3584}$:
\begin{equation}
    \mathbf{z}^{\mathrm{image}} = W_2\,\mathrm{GELU}(W_1\,\mathbf{f}^{\mathrm{patch}} + b_1) + b_2,
    \label{eq:projb}
\end{equation}
In Stage~1, the 576 image tokens are placed before the caption tokens to form the LLM input.
The training objective is next-token prediction on the caption only: image-token positions are masked, so they contribute no loss, and the projector is trained to make the image tokens useful for generating the caption that follows.

\subsection{Region Alignment (region bridge)}

The region bridge maps a bounding box to region tokens that describe its contents, following GPT4RoI~\citep{zhang2023gpt4roi} with the modifications below.
It comprises CoordConv, per-level projection, multi-head self-attention, RoI Align, multi-token attention pooling, and a final Region Feature Projector.
Starting from the same CLIP ViT used for the image bridge, intermediate transformer block features are taken, which represent early texture to higher-level object cues.
Following CoordConv~\citep{liu2018coordconv}, normalised spatial channels are concatenated onto each $24{\times}24$ feature map so every patch embedding position carries absolute spatial information.
Each depth is then passed through a linear layer and LayerNorm so content and coordinates are mixed and the four levels share a common feature space for multi-head self-attention.
Multi-head self-attention subsequently mixes them so that shallow texture and deeper semantics exchange information at the same patch embedding position.
RoI Align extracts a fixed-size window of per-level cropped features from each boxed region of these enriched per-level features.
Multi-token attention pooling then converts each window into $K$ tokens rather than a single pooled vector, and the Region Feature Projector maps those tokens into Qwen's embedding space.
Stage~2 trains the region bridge alone with CLIP and Qwen frozen. The paragraphs below detail each step.

\paragraph{Intermediate transformer block features}
GPT4RoI~\citep{zhang2023gpt4roi} also draws region features from multiple CLIP ViT-L/14 layers, selecting $\mathcal{L}=\{14,17,20,23\}$, which are the eleventh-, eighth-, fifth-, and second-to-last blocks.
That choice is also consistent with prior evidence that earlier ViT layers preserve texture and spatial layout while the last layer largely drops localisation~\citep{goldblum2022vitlearn}, that the penultimate CLIP layer typically outperforms the last for MLLMs~\citep{chen2025shallower}, and that CLIP's deepest layers favour image--text alignment over visual discriminability~\citep{zhou2025lhtclip}.
For each $\ell\in\mathcal{L}$, patch tokens (CLS discarded) form map $\mathbf{F}^{(\ell)}$.
The image bridge already uses only the final transformer block, which is the most image--text aligned but least localisation-preserving representation. The region bridge is therefore designed to read earlier and penultimate maps instead, so its region tokens retain finer spatial and object-level detail that the final block tends to lose.

\paragraph{CoordConv}
ViT positions are only implicit and may weaken after reshape/crop, so we append normalised spatial coordinate channels~\citep{liu2018coordconv} to each feature map to encode absolute spatial information.
Let $\mathbf{F}^{(\ell)}$ denote the level-$\ell$ visual map; after appending the two coordinate grids $\mathbf{x}_{\mathrm{grid}}$ and $\mathbf{y}_{\mathrm{grid}}$, the augmented map is denoted $\tilde{\mathbf{F}}^{(\ell)}$.
In words, this changes the feature vector at each patch embedding position from 1024 channels (visual content only) to 1026 channels (visual content plus absolute $x$/$y$ position).
CoordConv~\citep{liu2018coordconv} then mixes these coordinate channels with visual channels via a $1{\times}1$ channel-mixing operation. In ReVA, this mixing is implemented by the per-level linear layer in Eq.~\ref{eq:perlevel}.

\paragraph{Per-level projection}
After CoordConv, each level is a $1026$-d map that still lives in its own representational space: block~14 has seen far fewer self-attention blocks than block~23, so raw features are not directly comparable.
We therefore insert a learned projection before multi-head self-attention.
A linear layer is chosen because it plays two roles at once.
First, applied independently at each of the $576$ patch embedding positions it is mathematically equivalent to a $1{\times}1$ convolution over the $24{\times}24$ map, so it performs the CoordConv channel mixing after spatial channels are appended, mixing position with visual content.
Second, giving each level $\ell$ its own weights $W^{(\ell)}_{\mathrm{proj}}$ lets shallower and deeper maps learn different transforms into a common feature space, which is what makes multi-head self-attention meaningful.
Operating on the sequence layout rather than as \texttt{Conv2d} also keeps the module compatible with the attention that follows:
\begin{equation}
\begin{split}
    \mathbf{h}^{(\ell)}_{s}
    &= \mathrm{LN}^{(\ell)}\!\left(W^{(\ell)}_{\mathrm{proj}}\,\tilde{\mathbf{f}}^{(\ell)}_{s} + \mathbf{b}^{(\ell)}_{\mathrm{proj}}\right)
    \in \mathbb{R}^{1026}, \\
    &\qquad s \in \{1,\ldots,576\},\;\ell\in\mathcal{L}
\end{split}
\label{eq:perlevel}
\end{equation}
LayerNorm is chosen next because the four depths differ not only in semantics but in typical activation scale; without normalisation, multi-head self-attention would be biased by magnitude rather than by content.
$\mathrm{LN}^{(\ell)}$ places every level on a consistent scale so that the subsequent multi-head self-attention compares compatible vectors.

\paragraph{Multi-head self-attention}
With all four depths in a shared space, we apply multi-head self-attention across levels at every patch embedding position before RoI Align.
Stacking the projected vectors at patch embedding position $s$ gives a short sequence $\mathbf{H}_s\in\mathbb{R}^{4\times 1026}$; multi-head self-attention over this length-$4$ sequence mixes the four depths so that shallow texture and deeper semantics exchange information at the same patch embedding position:
\begin{equation}
    \hat{\mathbf{H}}_s = \mathrm{LN}\!\left(\mathbf{H}_s + \mathrm{MultiHeadAttn}(\mathbf{H}_s,\mathbf{H}_s,\mathbf{H}_s)\right)
    \label{eq:crosslevel}
\end{equation}
We use six heads and dropout $0.1$, with a residual connection and LayerNorm as in a standard transformer block \citep{vaswani2017attention}.
The $576$ patch embedding positions are processed independently (effective batch $B{\times}576$), so attention never mixes different patch embedding positions---only the four ViT depths at the same patch embedding position.
The attended sequences are then reshaped back to enriched per-level features $\hat{\mathbf{F}}^{(\ell)}$.

\paragraph{RoI Align and multi-token attention pooling}
RoI Align \citep{he2017maskrcnn} generally crops each proposed box to a fixed $n{\times}n$ feature window, preserving the features inside the box.
In ReVA we set $n{=}14$ (scale $24/336$) to avoid aggressive feature compression before pooling while keeping compute manageable, and to convert box coordinates from the 336-pixel image grid to the $24{\times}24$ grid of patch embedding positions before cropping:
\begin{equation}
    \mathbf{R}^{(\ell)}_{b} = \mathrm{RoIAlign}\!\left(\hat{\mathbf{F}}^{(\ell)},\,b;\,14{\times}14\right) \in \mathbb{R}^{1026\times 14\times 14},\;\ell\in\mathcal{L}
    \label{eq:roialign}
\end{equation}
Where $b$ is the bounding box.
During region-alignment training, GT boxes from the Stage~2 training corpora are used, and inference uses box proposals produced by the region detector.
These per-level cropped features are then pooled into $K$ tokens per level: each level has its own $K{=}16$ learnable queries that soft-attend over the $196$ cells.
Here each learnable query $k\in\{1,\ldots,K\}$ at depth $\ell$ produces one pooled vector $\mathbf{z}^{(\ell)}_{b,k}$:
\begin{equation}
\begin{split}
    \alpha^{(\ell)}_{b,k,i} &= \mathrm{softmax}_{i}\!\left(\frac{(\mathbf{q}^{(\ell)}_{k})^{\top} W^{(\ell)}_{K}\,\mathbf{r}^{(\ell)}_{b,i}}{\sqrt{1026}}\right),\;\ell\in\mathcal{L}
    \\
    \mathbf{z}^{(\ell)}_{b,k} &= \sum_{i=1}^{196} \alpha^{(\ell)}_{b,k,i}\, W^{(\ell)}_{V}\,\mathbf{r}^{(\ell)}_{b,i}
    \in \mathbb{R}^{1026},\;\ell\in\mathcal{L}
\end{split}
    \label{eq:mtap}
\end{equation}
\paragraph{Region Feature Projector}
Multi-token attention pooling yields $K$ vectors at each of the four ViT depths.
For every query index $k$, those four level-specific vectors are concatenated into one vector $\mathbf{u}_{b,k}\in\mathbb{R}^{4104}$; the Region Feature Projector---a two-layer MLP with LayerNorm, playing the same role as the Image Feature Projector---then maps it into Qwen space:
\begin{gather}
    \mathbf{u}_{b,k} = \mathrm{Concat}\!\left(\mathbf{z}^{(\ell)}_{b,k}\right) \in \mathbb{R}^{4104},\;\ell\in\mathcal{L}
    \label{eq:fuse}
    \\
    \mathbf{z}^{\mathrm{region}}_{b,k} = \mathrm{LN}_2\!\left(W^{A}_2\,\mathrm{LN}_1\!\left(\mathrm{GELU}\!\left(W^{A}_1\,\mathbf{u}_{b,k} + \mathbf{b}^{A}_1\right)\right) + \mathbf{b}^{A}_2\right)
    \label{eq:proja}
\end{gather}
Stage~2 prepends these to region descriptions (caption loss on text only) plus InfoNCE \citep{chen2020simple} ($\tau{=}0.07$, $\lambda_{\mathrm{itc}}{=}0.5$).
Each of $N$ boxes contributes $K$ tokens beside the $576$ image tokens.

\subsection{Region Detector}

At inference, RoI Align in the region bridge relies on automatic zero-shot bounding boxes to crop regional features.
RAM++ (Recognize Anything Model)~\citep{huang2023rampp} produces a tag set of what is present in the image (at most 20 tags); this tag set is united with spaCy nouns extracted from the question.
The union ensures that relevant nouns in the question are captured (via spaCy), while maintaining context about objects within the image (via RAM++).
Grounding~DINO~\citep{ren2024grounding} then localises each tag into at most one box.
Near-duplicate boxes are removed by NMS at IoU $\geq 0.95$, at most 20 boxes are kept, and if nothing is detected the stack falls back to a single whole-image box.
The boxes are used as crops for RoI Align only; they are not written into the prompt as text.
A sample token layout of the LLM prompt is shown below.
\begin{center}
\fbox{\begin{minipage}{0.94\columnwidth}
\vspace{0.3em}
\small\ttfamily
[576 image token embeddings]\\
{[}N $\times$ K region token embeddings{]}\\
\\
Answer the question using a single word or phrase.\\
\\
Question:\\
\textless question\textgreater\\
\\
Answer:
\vspace{0.3em}
\end{minipage}}
\end{center}

\section{Training Regime}
\label{sec:training}

Training is split into three stages so each bridge learns one alignment before the LLM combines them (Table~\ref{tab:training_stages}).
Stages~1--2 train the image bridge and the region bridge only, with CLIP and Qwen frozen; Stage~3 freezes the image bridge and the region bridge and updates only LoRA adapters in Qwen.

\begin{table*}[t]
\centering
\small
\setlength{\tabcolsep}{3pt}
\begin{tabular}{@{}c
  >{\raggedright\arraybackslash}p{0.22\textwidth}
  >{\raggedright\arraybackslash}p{0.30\textwidth}
  >{\raggedright\arraybackslash}p{0.36\textwidth}@{}}
\toprule
\textbf{Stage} & \textbf{Objective} & \textbf{Trainable module(s)} & \textbf{Training data} \\
\midrule
1 & Image--text alignment & Image bridge (Image Feature Projector) & LLaVA-Pretrain (LCS-558K)$^{\dagger}$ \\
2 & Region--description alignment & Region bridge (per-level projection, multi-head self-attention, multi-token attention pooling, Region Feature Projector) & COCO, RefCOCO/+/g, Visual Genome, GRIT-20M shard~0 ($\leq$250K rows)$^{\dagger}$ \\
3 & Grounded VQA instruction tuning & LoRA adapters (Qwen) & Visual7W (35\%), GQA (30\%), VQAv2 (15\%), VCR (10\%), A-OKVQA (10\%)$^{\dagger\ddagger}$ \\
\bottomrule
\end{tabular}
\caption{Overview of the three-stage ReVA training procedure.
$^{\dagger}$Stage~1: pHash (Hamming $\leq 4$) against COCO and Visual Genome; flagged pairs treated as false positives; full 558{,}128-image pool retained.
Stage~2: pHash (Hamming $\leq 4$) against POPE, VQAv2 test2015, MMBench, and SEED-Bench.
Stage~3: hard-ID exclusion then pHash against the same eval image union.
$^{\ddagger}$Stage~3 region crops use up to 20 boxes per image across all sources: VQAv2 and A-OKVQA use COCO train2014 ground-truth boxes; GQA, Visual7W, and VCR use their own box annotations.}
\label{tab:training_stages}
\end{table*}

\begin{table*}[t]
\centering
\footnotesize
\setlength{\tabcolsep}{2.5pt}
\setlength{\extrarowheight}{0pt}
\begin{tabular}{@{}l
  >{\raggedright\arraybackslash}p{0.24\textwidth}
  >{\raggedright\arraybackslash}p{0.24\textwidth}
  >{\raggedright\arraybackslash}p{0.24\textwidth}@{}}
\toprule
\textbf{Hyperparameter} & \textbf{Stage 1} & \textbf{Stage 2} & \textbf{Stage 3} \\
\midrule
Trainable module(s) &
Image bridge &
Region bridge &
Qwen LoRA adapters ($q/k/v/o$, gate/up/down) \\
Epochs / optimiser steps & 1 epoch & 3 epochs & 10{,}000 steps \\
Per-device / micro-batch size & 16 & 16 & 1 \\
Gradient accumulation & 16 & 8 & 16 \\
Effective batch size & 256 & 128 & 16 \\
Peak learning rate & $10^{-3}$ & $7{\times}10^{-5}$ & $10^{-4}$ \\
Warmup ratio & 3\% & 6\% & 3\% \\
Loss objective &
Cross-entropy on caption tokens &
$L_{\mathrm{cap}}+\lambda_{\mathrm{itc}}L_{\mathrm{itc}}$ (Eq.~\ref{eq:stage2loss}) &
Cross-entropy on answer tokens \\
$\lambda_{\mathrm{itc}}$ / $\tau$ & --- & 0.5 / 0.07 & --- \\
Max target tokens & 128 (caption) & 64 (description) & 32 (8 on A-OKVQA) \\
Region tokens per box ($K$) & --- & 16 & 16 \\
Max boxes per image & --- & GT boxes in Stage~2 corpora & 20 \\
LoRA rank $r$ / $\alpha$ / dropout & --- & --- & 16 / 32 / 0.05 \\
Multi-head self-attention heads & --- & 6 & --- \\
Attention dropout & --- & 0.1 & --- \\
\bottomrule
\end{tabular}
\captionsetup{justification=centering,singlelinecheck=true}
\caption{Principal hyperparameters for the three-stage ReVA training pipeline.}
\captionsetup{justification=raggedright,singlelinecheck=false}
\label{tab:hyperparams}
\end{table*}

\subsection{Stage 1: Image Alignment}
\label{sec:stage1}

Stage~1 aligns the image bridge (Image Feature Projector) to Qwen on whole images.
We ran pHash \citep{zauner2010implementation} (Hamming $\leq 4$) on LLaVA-Pretrain (LCS-558K)~\citep{liu2024visual} against COCO \citep{lin2014microsoft} and Visual Genome \citep{krishna2017visualgenome}.
The flagged pairs were false positives, so we retained the full 558{,}128-image pool.

Only the image bridge ($\sim$16.5M parameters) is trained; CLIP and Qwen stay frozen, with gradient checkpointing on Qwen (Table~\ref{tab:hyperparams}).

\subsection{Stage 2: Region Alignment}
\label{sec:stage2}

Stage~2 trains the region bridge to describe box contents using decontaminated region--text pairs: COCO names \citep{lin2014microsoft}, RefCOCO/+/g expressions \citep{yu2016refcoco}, Visual Genome attributes \citep{krishna2017visualgenome} (capped at 15 regions/image), and GRIT-20M \citep{peng2023kosmos2} phrases.
For this stage, training uses GT box proposals, while inference later uses Grounding~DINO box proposals \citep{ren2024grounding}.
We use only GRIT metadata shard~0, keep rows with grounded boxes and CLIP ViT-L/14 \citep{radford2021learning} similarity $\geq 0.30$, and cap at 250{,}000 rows before download.
pHash (Hamming $\leq 4$) removes overlap with POPE, VQAv2 test2015, MMBench, and SEED-Bench.
For each (image, box, description) we minimized
\begin{equation}
    L = L_{\mathrm{cap}} + \lambda_{\mathrm{itc}}\,L_{\mathrm{itc}}
    \label{eq:stage2loss}
\end{equation}
i.e.\ teach fluent box captions ($L_{\mathrm{cap}}$) while also pulling matching region/text pairs together in embedding space and pushing mismatches apart ($L_{\mathrm{itc}}$, InfoNCE~\citep{chen2020simple} at $\tau{=}0.07$).
$K$ region tokens are prepended; only description positions receive loss (region-token positions are masked).
We set $\lambda_{\mathrm{itc}}{=}0.5$.
The combined region--text pool is much larger for Visual Genome and COCO than for RefCOCO and GRIT, so a uniform shuffle would train mostly on the two large sources.
We therefore draw each training batch so that the four sources contribute roughly equally, even though that means repeating some RefCOCO and GRIT pairs and skipping some Visual Genome and COCO pairs in a given epoch.

\subsection{Stage 3: LoRA Instruction Tuning}
\label{sec:stage3}

After the image bridge and the region bridge are trained, Stage~3 teaches Qwen to use image and region tokens for VQA.
We use a Stage~3 pool stored after decontamination, built from Visual7W \citep{yu2016visual7w}, GQA \citep{hudson2019gqa}, VQAv2 \citep{goyal2017making}, VCR \citep{zellers2019vcr}, and A-OKVQA \citep{schwenk2022aokvqa}, with up to 20 boxes per image: VQAv2 and A-OKVQA use COCO train2014 ground-truth boxes \citep{lin2014microsoft}, while GQA, Visual7W, and VCR use their own box annotations. Rather than sampling uniformly from the decontaminated pool---which is heavily skewed toward GQA and VQAv2---we target a fixed mix of Visual7W, GQA, VQAv2, VCR, and A-OKVQA at 35\%, 30\%, 15\%, 10\%, and 10\%, ensuring adequate exposure to pointing (Visual7W), spatial awareness (GQA), open-ended perception (VQAv2), commonsense reasoning (VCR), and MCQ-style answering (A-OKVQA).

LoRA \citep{hu2022lora} updates Qwen attention and MLP layers; the image bridge and the region bridge stay frozen.
Training runs for 10{,}000 optimiser steps with a weighted sampler that applies those source weights (Table~\ref{tab:hyperparams}).

\section{Experimental Setup}
\label{sec:setup}

\subsection{Benchmarks and Metrics}
\label{sec:metrics}

We report four primary Stage~3 suites, chosen because they stress complementary failure modes of whole-image MLLMs and are widely used in the literature we compare against.

\textbf{VQAv2}~\citep{goyal2017making} (test-dev / test-standard) measures open-ended short-answer accuracy with the official soft-scoring protocol.
Each question has ten human answers; a prediction that matches at least three annotators receives full credit, while rarer matches receive partial credit.
We follow the official punctuation and digit/article normalisation of \texttt{vqaEval.py}.
Gains here indicate that region tokens do not hurt broad question answering.

\textbf{MMBench}~\citep{li2024mmbench} (English) is a multiple-choice suite covering perception and reasoning skills.
Accuracy under CircularEval (shuffled options) reduces lucky guessing and tests whether the model follows the intended choice, not just fluent text.

\textbf{POPE}~\citep{li2023pope} probes object hallucination with yes/no questions on COCO in random, popular, and adversarial splits.
We report per-split and mean F1 (Table~\ref{tab:pope}): precision/recall matter because models that over-answer ``yes'' can look accurate while hallucinating.
Adversarial POPE is especially relevant to ReVA's claim that local evidence should suppress false object claims.

\textbf{SEED-Bench}~\citep{li2023seed} image split (SEED-Image; evaluation dimensions~1--9) is a broad multiple-choice comprehension benchmark.
We report image-only accuracy, not SEED~All, because ReVA is an image VQA system and our evaluator excludes video questions.
Overall SEED-Image accuracy summarises whether region grounding preserves general multimodal competence.

Stage~2 and Stage~3 training pools are pHash-decontaminated against the reported eval image sets; Stage~3 also drops exact question/image ID overlaps.
All reported evaluations use greedy decoding.
Where instance-level predictions are available, we additionally report two-sided paired $t$-tests and McNemar tests on matched questions between the image-token baseline and the full system (Tables~\ref{tab:paired_ttest} and~\ref{tab:mcnemar}).

\subsection{Implementation}

Experiments use PyTorch, Hugging Face Transformers, and PEFT LoRA.
Inference proposals: Grounding~DINO (Swin-T OGC) and RAM++ (Swin-L) on NVIDIA A100 GPUs (bf16).

\section{Results and Ablations}
\label{sec:results}

\subsection{Image Backbone Ablation}
\label{sec:ablations_backbone}

Stage~1 trains only the image bridge, so the backbone decision is independent of the region bridge and of LoRA.
Table~\ref{tab:backbone} compares frozen encoders under that setting; all runs use Qwen2.5-7B-Instruct, and $^{*}$ marks Stage~1 zero-shot with frozen CLIP/Qwen.
Raising CLIP from 224 to 336\,px improves GQA val by 3.90 points and VQAv2 val by 3.75, consistent with a $24{\times}24$ (576-token) grid versus $16{\times}16$ (256).
DINOv2-ViT-L/14 \citep{oquab2024dinov2} trails CLIP-ViT-L/14-336.
We therefore fix CLIP ViT-L/14 at 336\,px for all later stages.

\begin{table*}[!t]
\centering
\small
\begin{tabular}{@{}llccccc@{}}
\toprule
\textbf{Model} & \textbf{Backbone} & \textbf{GQA val} & \textbf{VQAv2 val} & \textbf{GQA test-dev} & \textbf{VQAv2 test-dev} \\
\midrule
ReVA (image tokens)$^{*}$ & CLIP-L/14 224 & 39.06 & 37.24 & -- & -- \\
ReVA (image tokens)$^{*}$ & CLIP-L/14 336 & \textbf{42.96} & \textbf{40.99} & \textbf{39.74} & \textbf{40.22} \\
ReVA (image tokens)$^{*}$ & DINOv2-L/14 & 37.84 & 35.81 & -- & -- \\
\bottomrule
\end{tabular}
\caption{Image backbone ablation (Stage~1, image alignment only).
All models use Qwen2.5-7B-Instruct.
$^{*}$Stage~1 zero-shot: frozen CLIP/Qwen; only the image bridge (Image Feature Projector) is trained.}
\label{tab:backbone}
\end{table*}

\subsection{Image-Token Baseline vs.\ Full System}
\label{sec:ablations_full}
\label{sec:sota}

The central comparison is whether region tokens help after LoRA teaches Qwen to use them.
The \emph{image-token} baseline keeps the Stage~1 image bridge and Stage~3 LoRA \citep{hu2022lora}, but omits region tokens; the \emph{full} system adds the Stage~2 region bridge and, at inference, RAM++ \citep{huang2023rampp}/Grounding~DINO \citep{ren2024grounding} boxes.
Table~\ref{tab:sota} reports that comparison against published MLLMs on VQAv2, MMBench, and SEED-Image; Table~\ref{tab:pope} isolates object hallucination on POPE.
All accuracy values use the protocols in Section~\ref{sec:metrics}: VQAv2 soft accuracy (test-dev / test-standard), MMBench EN \textit{dev} under CircularEval, and SEED-Image (dims~1--9), not SEED~All.
$\Delta$ is the difference between ReVA with image and region tokens and ReVA with image tokens only.

\begin{table*}[!t]
\centering
\footnotesize
\setlength{\tabcolsep}{3pt}
\begin{tabular}{@{}llcccc@{}}
\toprule
\textbf{Model} & \textbf{LLM} & \textbf{VQAv2 test-dev} & \textbf{VQAv2 test-std} & \textbf{MMBench} & \textbf{SEED-Bench} \\
\midrule
BLIP-2 & OPT-6.7B & 82.19 & 82.30 & --- & --- \\
InstructBLIP$^{*}$ & Vicuna-7B & --- & --- & 36.0$^\dagger$ & 58.8$^{\sharp}$ \\
LLaVA-1.5 & Vicuna-7B & 78.5 & --- & 64.3$^{*}$ & 66.1$^{\sharp}$ \\
LLaVA-1.5 & Vicuna-13B & 80.0 & --- & 67.7$^{*}$ & 68.2$^{\sharp}$ \\
Shikra & Vicuna-7B & 77.36 & 77.51 & --- & --- \\
Ferret-v2 & Vicuna-7B & 81.5 & --- & --- & --- \\
Qwen-VL$^{*}$ & Qwen-7B & 79.5 & --- & 38.2$^\dagger$ & 62.3$^{\sharp}$ \\
Qwen-VL-Chat$^{*}$ & Qwen-7B & 78.2 & --- & 60.6$^\dagger$ & 65.4$^{\sharp}$ \\
\midrule
GPT-4V$^{*}$ & --- & --- & --- & 75.0$^\S$ & 71.6$^\S$ \\
Gemini Pro$^{*}$ & --- & --- & --- & 75.2$^\S$ & 70.7$^\S$ \\
GPT-4o$^{*}$ & --- & --- & --- & 83.4$^\ddagger$ & --- \\
Claude 3.5 Sonnet$^{*}$ & --- & --- & --- & 79.7$^\ddagger$ & --- \\
DeepSeek-VL$^{*}$ & DeepSeek-7B & --- & --- & 73.2$^\S$ & 70.4$^\S$ \\
\midrule
\textbf{ReVA (image tokens)} & Qwen2.5-7B & 72.89 & 73.14 & 68.2$^{*}$ & 66.57$^{*}$ \\
\textbf{ReVA (image + region)} & Qwen2.5-7B & 73.28 & 73.79 & 67.1$^{*}$ & 66.75$^{*}$ \\
$\Delta$ & --- & $+0.39$ & $+0.65$ & $-1.1$ & $+0.18$ \\
\bottomrule
\end{tabular}
\caption{Comparison with state-of-the-art multimodal LLMs on VQAv2, MMBench, and SEED-Bench.
Values are accuracy (\%); SEED-Bench is image-only (dims~1--9), not SEED~All.
$\Delta$ is the difference between ReVA with image and region tokens and ReVA with image tokens only.
$^{*}$Benchmark zero-shot; unmarked VQAv2 rows observe VQAv2 train images.
$^\dagger$LLaVA-1.5 re-evaluation~\citep{liu2023llava}; $^\S$DeepSeek-VL~\citep{lu2024deepseekvl}; $^\ddagger$DeepSeek-VL2~\citep{wu2024deepseekvl2}; $^{\sharp}$SEED-Bench (\texttt{img}) from LLaVA-1.5~\citep{liu2023llava} Table~4.}
\label{tab:sota}
\end{table*}

\begin{table*}[!t]
\centering
\small
\setlength{\tabcolsep}{3pt}
\begin{tabular}{@{}llcccc@{}}
\toprule
\multirow{2}{*}{\textbf{Model}} & \multirow{2}{*}{\textbf{LLM}} & \multicolumn{4}{c}{\textbf{POPE}} \\
\cmidrule(lr){3-6}
& & Random & Popular & Adversarial & Mean \\
\midrule
LLaVA-1.5$^{*}$ & Vicuna-7B & 87.3 & 86.1 & 84.2 & 85.9 \\
LLaVA-1.5$^{*}$ & Vicuna-13B & 87.1 & 86.2 & 84.5 & 85.9 \\
Shikra$^{*}$ & Vicuna-7B & 86.19 & 83.16 & 82.49 & 83.9 \\
Ferret$^{*}$ & Vicuna-7B & 89.76 & 84.21 & 82.00 & 85.3 \\
Ferret-v2$^{*}$ & Vicuna-7B & --- & --- & --- & 87.8 \\
DeepSeek-VL$^{*}$ & DeepSeek-7B & --- & --- & --- & 88.1$^\S$ \\
\midrule
\textbf{ReVA (image tokens)}$^{*}$ & Qwen2.5-7B & 81.99 & 81.26 & 80.18 & 81.14 \\
\textbf{ReVA (image + region)}$^{*}$ & Qwen2.5-7B & 84.10 & 82.71 & 81.75 & 82.85 \\
$\Delta$ & --- & $+2.11$ & $+1.45$ & $+1.57$ & $+1.71$ \\
\bottomrule
\end{tabular}
\caption{POPE F1 (\%) on COCO (random / popular / adversarial; Mean is the unweighted average).
$^{*}$Benchmark zero-shot; $^\S$from DeepSeek-VL~\citep{lu2024deepseekvl}.
$\Delta$ is the difference between ReVA with image and region tokens and ReVA with image tokens only.
Baselines from~\citep{liu2023llava,you2024ferret,zhang2024ferretv2,lu2024deepseekvl}.}
\label{tab:pope}
\end{table*}

On VQAv2, the full system raises soft accuracy from 72.89 to 73.28 on test-dev ($+0.39$) and from 73.14 to 73.79 on test-standard ($+0.65$).
Region tokens leave broad multiple-choice competence essentially unchanged: MMBench drops by $1.1$ points while SEED-Image rises by $0.18$.
The MMBench dip is a skill-level trade-off rather than a uniform regression: localisation and attribute comparison improve, while coarse perception and relation reasoning fall (Section~\ref{sec:discussion}).
The clearer signal is POPE (Table~\ref{tab:pope}), where the full system improves mean F1 by $+1.71$ over the image-token baseline (random $+2.11$, popular $+1.45$, adversarial $+1.57$), consistent with local evidence helping suppress in-image object hallucination.
Section~\ref{sec:paired_ttest} tests that claim with instance-level paired $t$-tests and McNemar tests.
Against open-source peers, ReVA remains below stronger VQAv2 numbers from models that train on VQAv2 images more aggressively, but the image-token vs.\ full contrast is the relevant control for the region-token claim.
ReVA rows use greedy decoding with option-letter matching; MMBench is scored with VLMEvalKit \citep{duan2024vlmevalkit} CircularEval, and ReVA's 66.57 is SEED-Image from our evaluator (\texttt{data\_type=image}, \texttt{question\_type\_id}$\in\{1,\ldots,9\}$).

\subsection{Statistical Significance}
\label{sec:paired_ttest}

\begin{figure*}[!p]
\centering
\vspace*{-1.5em}%
\includegraphics[width=\textwidth]{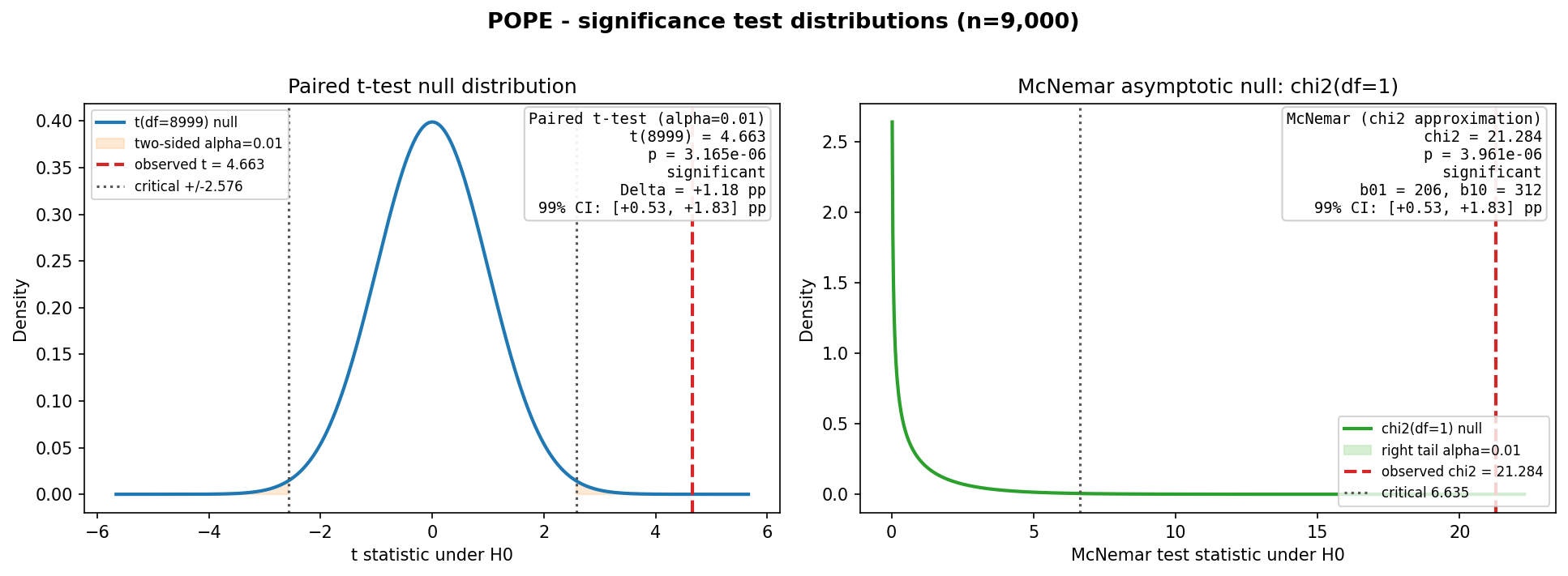}\\[0.45em]
\includegraphics[width=\textwidth]{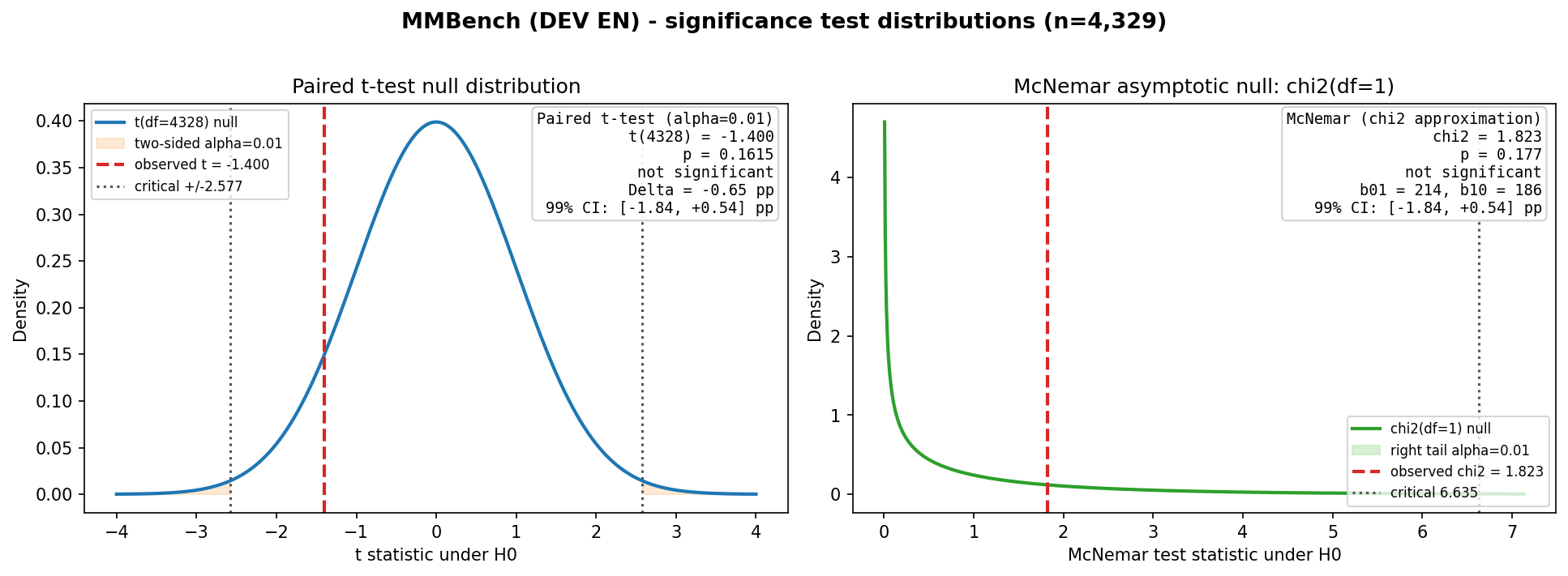}\\[0.45em]
\includegraphics[width=\textwidth]{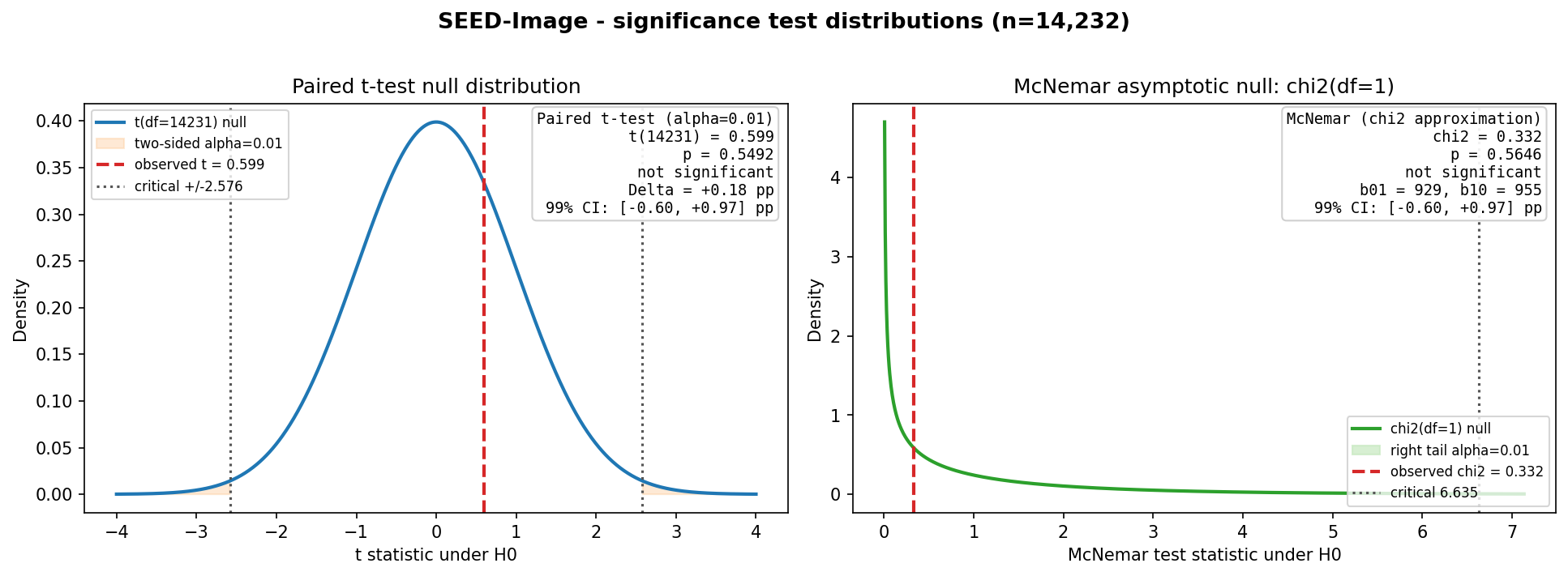}%
\refstepcounter{figure}\label{fig:paired_tests}
\end{figure*}

To test whether region tokens improve localised grounding and reduce hallucination, we compare the image-token baseline and the full ReVA system on matched questions using binary correctness (Tables~\ref{tab:paired_ttest} and~\ref{tab:mcnemar}).
We report a two-sided paired $t$-test ($H_0{:}\,\mathrm{mean}(d){=}0$) and McNemar's test on discordant pairs ($H_0{:}\,P(b_{01}){=}P(b_{10})$), both at $\alpha{=}0.01$.
Here $b_{01}$ counts questions the base got right and the final system got wrong (region tokens flipped a correct answer), while $b_{10}$ counts questions the base got wrong and the final system got right (region tokens fixed an incorrect answer).
POPE is pooled across random/popular/adversarial as accuracy (not the F1 in Table~\ref{tab:pope}); MMBench uses per-question JSONL correctness rather than the CircularEval Overall in Table~\ref{tab:sota}; SEED-Image uses dims~1--9.
VQAv2 is excluded from these paired tests because labeled per-question JSONL dumps were unavailable; Table~\ref{tab:paired_ttest} therefore lists EvalAI overall soft scores only.
Fig.~\ref{fig:paired_tests} shows the corresponding null distributions.

\begin{table}[t]
\centering
\scriptsize
\setlength{\tabcolsep}{3pt}
\begin{tabular}{@{}lrrrrrc@{}}
\toprule
\textbf{Benchmark} & \textbf{$n$} & \textbf{Base\%} & \textbf{Final\%} & \textbf{$\Delta$} & \textbf{$p_t$} & \textbf{Sig?} \\
\midrule
POPE & 9{,}000 & 83.63 & 84.81 & $+1.18$ & $3.165{\times}10^{-6}$ & Yes$^{\dagger}$ \\
MMBench & 4{,}329 & 77.96 & 77.32 & $-0.65$ & $0.1615$ & No \\
SEED-Image & 14{,}232 & 66.57 & 66.75 & $+0.18$ & $0.5492$ & No \\
VQAv2 & --- & 72.89 & 73.28 & $+0.39$ & --- & --- \\
\bottomrule
\end{tabular}
\caption{Paired $t$-test (Base: image tokens; Final: image + region).
$^{\dagger}$Significant at $\alpha{=}0.01$.}
\label{tab:paired_ttest}
\end{table}

\begin{table}[t]
\centering
\scriptsize
\setlength{\tabcolsep}{3pt}
\begin{tabular}{@{}lrrrrc@{}}
\toprule
\textbf{Benchmark} & \textbf{$n$} & \textbf{$b_{01}$} & \textbf{$b_{10}$} & \textbf{$p_{\mathrm{McN}}$} & \textbf{Sig?} \\
\midrule
POPE & 9{,}000 & 206 & 312 & $3.961{\times}10^{-6}$ & Yes$^{\dagger}$ \\
MMBench & 4{,}329 & 214 & 186 & $0.177$ & No \\
SEED-Image & 14{,}232 & 929 & 955 & $0.5646$ & No \\
\bottomrule
\end{tabular}
\caption{McNemar test (Base: image tokens; Final: image + region) on discordant pairs.
$b_{01}$: base correct, final incorrect (region tokens flipped a previously correct answer);
$b_{10}$: base incorrect, final correct (region tokens fixed a previously wrong answer).
$^{\dagger}$Significant at $\alpha{=}0.01$.}
\label{tab:mcnemar}
\end{table}

On POPE accuracy, the full system gains $+1.18$ percentage points ($p_t{=}3.165{\times}10^{-6}$; McNemar $p{=}3.961{\times}10^{-6}$), significant at $\alpha{=}0.01$ under both tests.
MMBench and SEED-Image show no significant change ($p_t{=}0.1615$ and $0.5492$; McNemar $p{=}0.177$ and $0.5646$).
Overall, region tokens help most on object-hallucination probes, where local evidence matters, and do not significantly change broad multiple-choice scores on MMBench or SEED-Image.

\section{Discussion}
\label{sec:discussion}

\paragraph{Decontamination and data scale}
We ran pHash (Hamming $\leq 4$) on LCS-558K against COCO and Visual Genome.
The flagged pairs were false positives, so we retained the full 558{,}128-image pool (Table~\ref{tab:decontam}, Kept $=$ Original).
Stage~2 training images were decontaminated against POPE, VQAv2, MMBench, and SEED-Bench using pHash.
Stage~3 fine-tuning images were decontaminated against POPE, VQAv2, MMBench, and SEED-Bench using hard-ID exclusion then pHash.
Although the risk of contamination is lowered for Stages~2--3, residual overlap with evaluation images remains possible; Table~\ref{tab:decontam} reports how much was removed, not that the remaining pools are fully clean.

\begin{table}[t]
\centering
\small
\setlength{\tabcolsep}{3.5pt}
\begin{tabular}{@{}lrrr@{}}
\toprule
\textbf{Stage} & \textbf{Original} & \textbf{Kept} & \textbf{Removed} \\
\midrule
1 (LCS-558K) & 558{,}128 & 558{,}128 & 0.00\% \\
2 (region corpora) & 2{,}951{,}772 & 2{,}940{,}403 & 0.39\% \\
3 (VQA pool) & 1{,}704{,}599 & 1{,}699{,}447 & 0.30\% \\
\bottomrule
\end{tabular}
\caption{Training-pool size after decontamination.}
\label{tab:decontam}
\end{table}

\paragraph{MMBench skill trade-off}
\label{sec:mmbench_tradeoff}
The CircularEval Overall drop of $1.1$ points on MMBench (Table~\ref{tab:sota}) is not uniform across skills.
Relative to the image-token baseline, the full system gains on object localisation ($+8.65$), attribute comparison ($+6.82$), identity reasoning ($+4.45$), fine-grained perception, and attribute reasoning---skills that benefit from cropped local evidence.
It reduces performance on coarse perception ($-4.73$), relation reasoning ($-5.21$), OCR ($-5.13$), attribute recognition ($-6.76$), and on image-quality judgement ($-24.53$).
This fits a capacity trade-off in which region tokens use prefix space for local object evidence at the expense of holistic scene and relation cues.
The Overall drop is not significant under the paired $t$-test or McNemar test (Tables~\ref{tab:paired_ttest} and~\ref{tab:mcnemar}), so it is better read as a redistribution of skill than a reliable regression.

\section{Future Work}
\label{sec:future}

\paragraph{Train--test box gap}
Replacing GT boxes with Grounding~DINO proposals at training time---or mixing clean and noisy boxes---would shrink the mismatch that appears at inference.
A light region reranker (score boxes by question relevance before packing the prefix) is a practical middle ground: keep at most $N$ regions without forcing the LLM to attend to every detector proposal.

\paragraph{Architecture and compute}
The region bridge adds up to $20{\times}K$ tokens on top of the 576 image tokens, plus RAM++/Grounding~DINO latency, so future variants should prune low-confidence or overlapping boxes, try a smaller $K$, or distill the region bridge into a lighter encoder.
A CoordConv on/off ablation would isolate how much explicit spatial channels matter once RoI Align already crops local features.
Swapping the frozen vision encoder under the same two-projector recipe would test whether the gains are backbone-specific.

\paragraph{Broader grounded tasks}
The current version of ReVA is trained on short-phrase VQA.
Further work can extend it to open-ended long answers, multi-turn referring dialogue, grounded captioning, and harder visual reasoning that needs several local clues in sequence.

\paragraph{Multi-seed replication}
Tables~\ref{tab:paired_ttest}--\ref{tab:mcnemar} compare one checkpoint pair, so the tests measure question-level reliability rather than stability across training runs.
Retraining Stages~1--3 with several seeds would show whether the POPE gain, and the non-significant MMBench and SEED-Image differences, hold more generally.

\vspace{-1.25ex}
\section{Conclusion}
\label{sec:conclusion}

This work asks whether ReVA answers visual questions more reliably when the language model receives both image tokens and explicit region tokens.
On POPE, the full system improves mean F1 on every split over the image-token baseline (Table~\ref{tab:pope}).
On VQAv2, soft accuracy rises on both test-dev and test-standard, completing the image-token versus image + region comparison alongside POPE, MMBench, and SEED-Image (Table~\ref{tab:sota}).
These results suggest that ReVA answers visual questions more reliably when it receives both the whole image and explicit region tokens from multi-level RoI-cropped ViT features, improving object-level grounding and reducing hallucination while largely preserving general VQA capability relative to a whole-image encoding alone.

\section*{Acknowledgment}

I thank Prof.\ Shalom Lappin for his supervision, guidance, and feedback throughout this project.
I am also grateful to the School of Electronic Engineering and Computer Science at Queen Mary University of London, and to the comp-teach team that made this work possible.

\bibliographystyle{agsm}
\bibliography{references}

\makeatletter
\setlength{\@fptop}{0pt}
\setlength{\@fpbot}{0pt plus 1fil}
\makeatother
\begin{figure*}[!p]
\centering
\vspace*{-1em}%
\adjustbox{max width=\textwidth, max height=0.90\textheight, keepaspectratio, center}{%
\begin{tabular}{@{}c@{\hspace{5pt}}c@{}}
\includegraphics[height=0.44\textheight,valign=t]{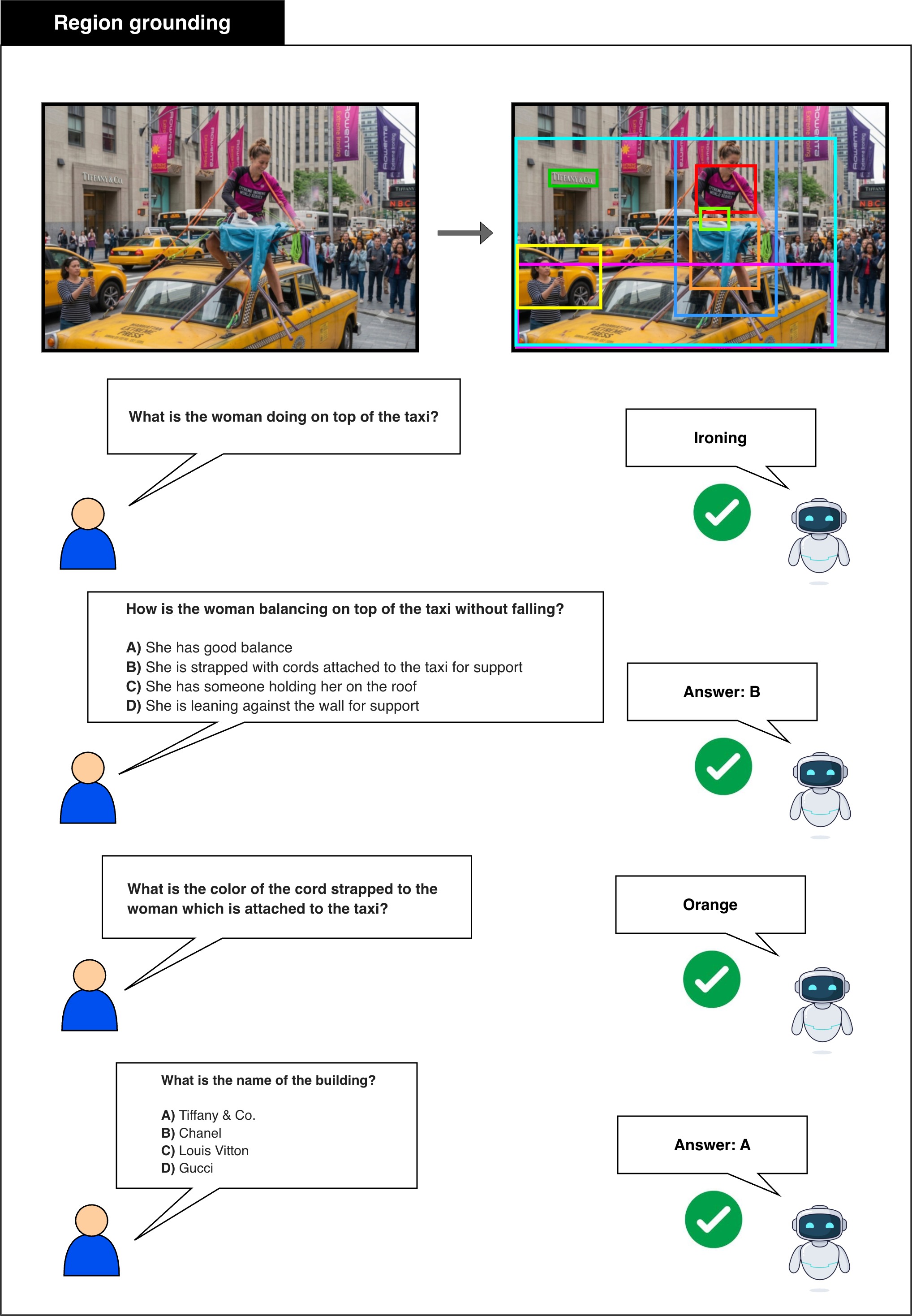}%
\refstepcounter{figure}\label{fig:example_queries} &
\includegraphics[height=0.44\textheight,valign=t]{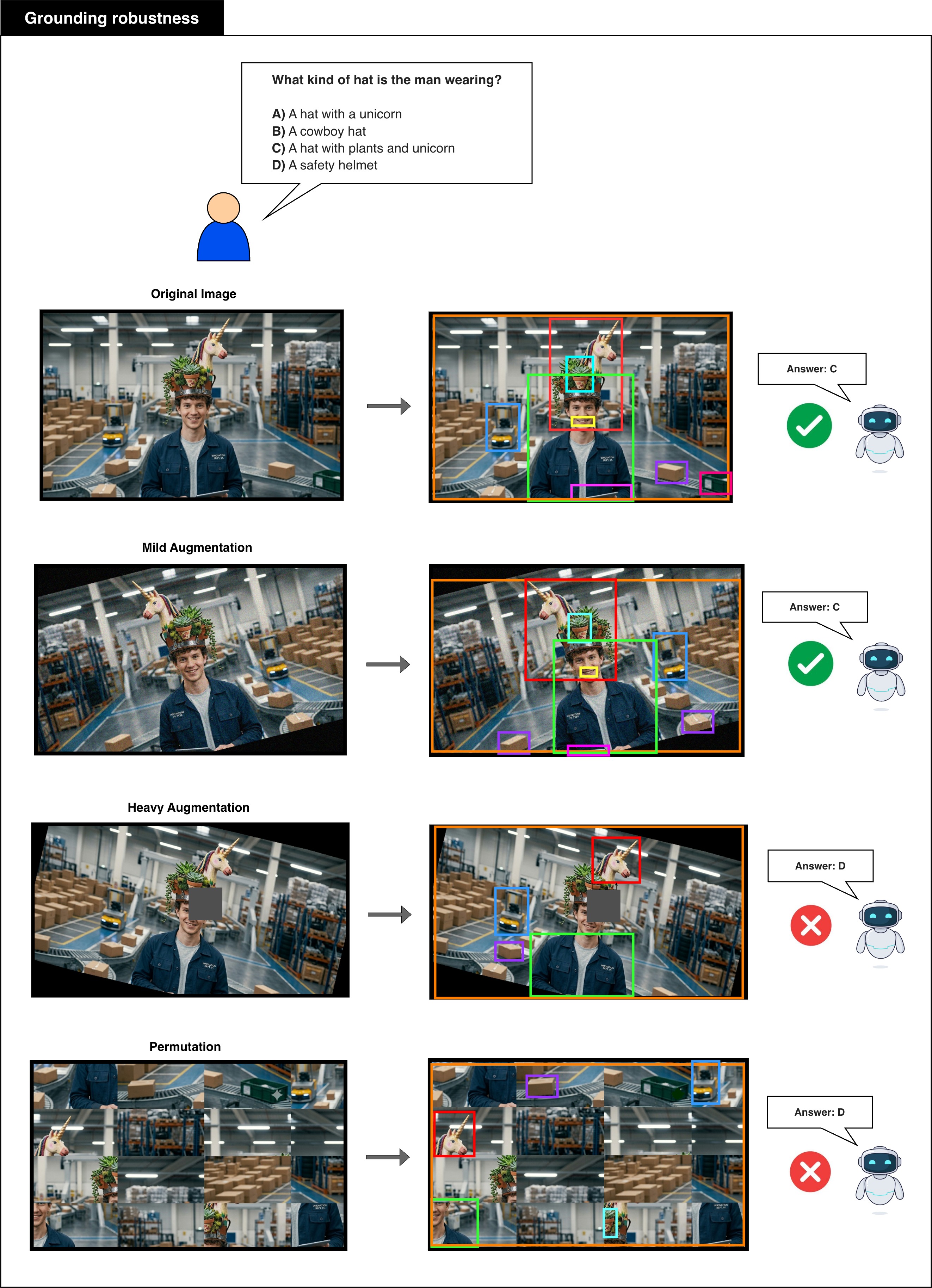}%
\refstepcounter{figure}\label{fig:grounding} \\[0.4em]
\includegraphics[height=0.58\textheight,valign=t]{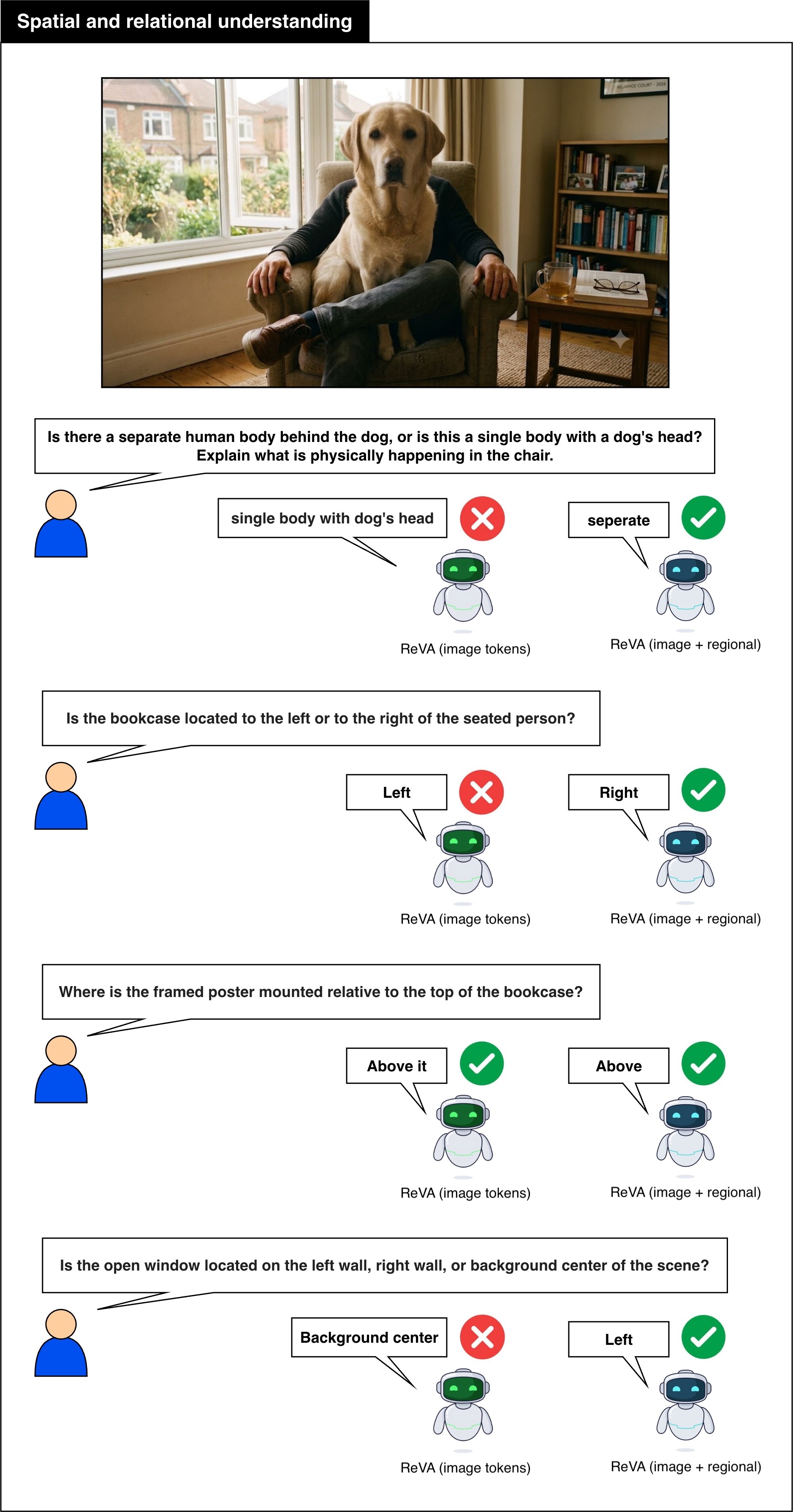}%
\refstepcounter{figure}\label{fig:example_spatial} &
\includegraphics[height=0.66\textheight,valign=t]{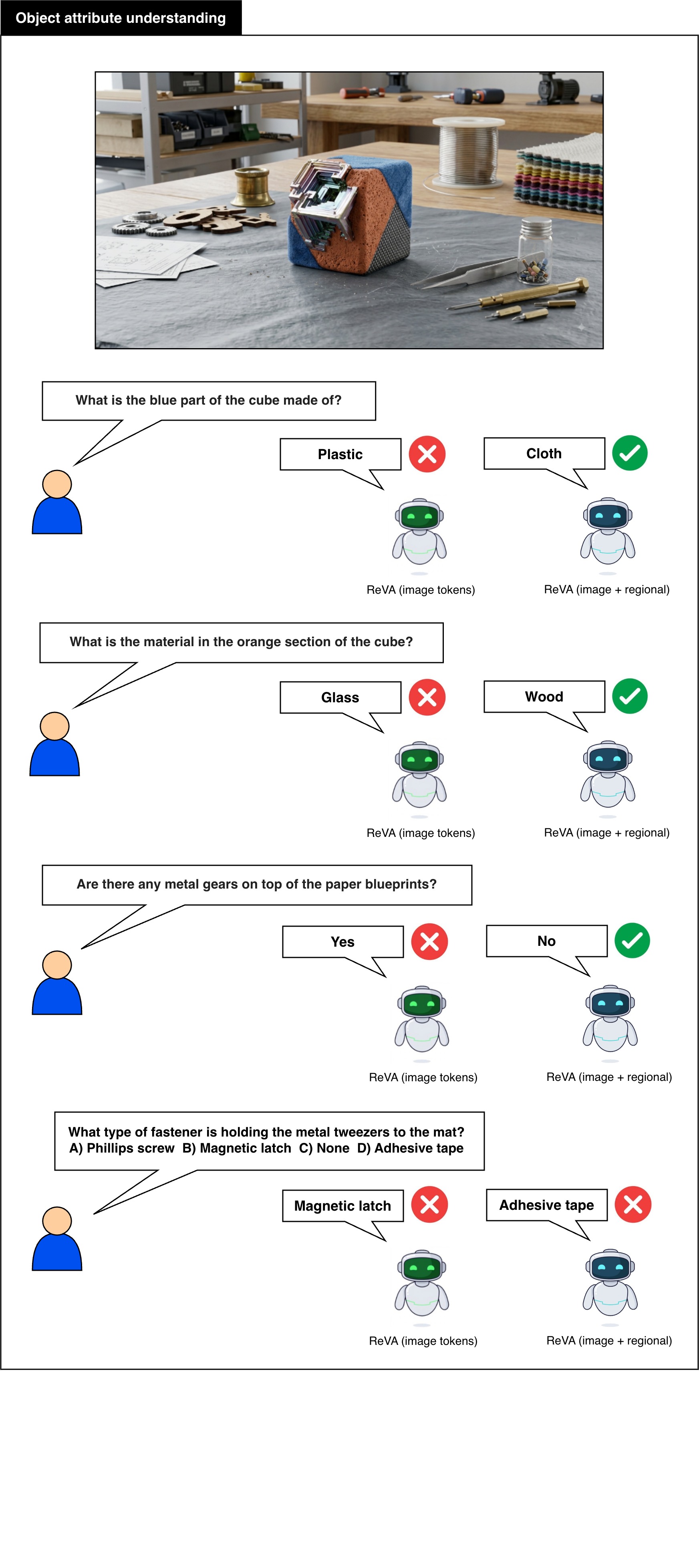}%
\refstepcounter{figure}\label{fig:example_attributes}
\end{tabular}}
\end{figure*}

\end{document}